\documentclass{article}
\usepackage{spconf,amsmath,graphicx}

\title{EM-NET: CENTERLINE-AWARE MITOCHONDRIA SEGMENTATION IN EM IMAGES VIA HIERARCHICAL VIEW-ENSEMBLE CONVOLUTIONAL NETWORK}
\name{Zhimin Yuan, Jiajin Yi, Zhengrong Luo, Zhongdao Jia, and Jialin Peng$^*$}
\address{College of Computer Science and Technology, Huaqiao University, Xiamen, China\\
*2004pjl@163.com\thanks{*Corresponding author. J. Peng was supported by NSFC (11771160), and  STPF (2019H0016).}}
\begin{document}
%
\maketitle
\begin{abstract}
Although deep fully convolutional networks have achieved astonishing performance for mitochondria segmentation from electron microscopy (EM) images, they still produce coarse segmentations with  discontinuities and false positives. Besides, the need for labor intensive annotations of large 3D volumes and huge memory overhead by 3D models are also major limitations. To address these problems, we introduce a multi-task network named EM-Net, which includes an auxiliary centerline detection task to account for shape information of mitochondria represented by centerline.  Therefore, the centerline detection sub-network is able to enhance the accuracy and robustness of segmentation task, especially when only a small set of annotated data are available. To achieve a light-weight 3D network, we introduce a novel \textit{hierarchical view-ensemble convolution} module to reduce number of parameters, and facilitate multi-view information aggregation. Validations on public benchmark showed state-of-the-art performance by EM-Net. Even with significantly reduced training data, our method still showed quite promising results.
\end{abstract}

\begin{figure*}
\begin{minipage}[b]{1.0\linewidth}
  \centerline{\includegraphics[width=0.75\textwidth]{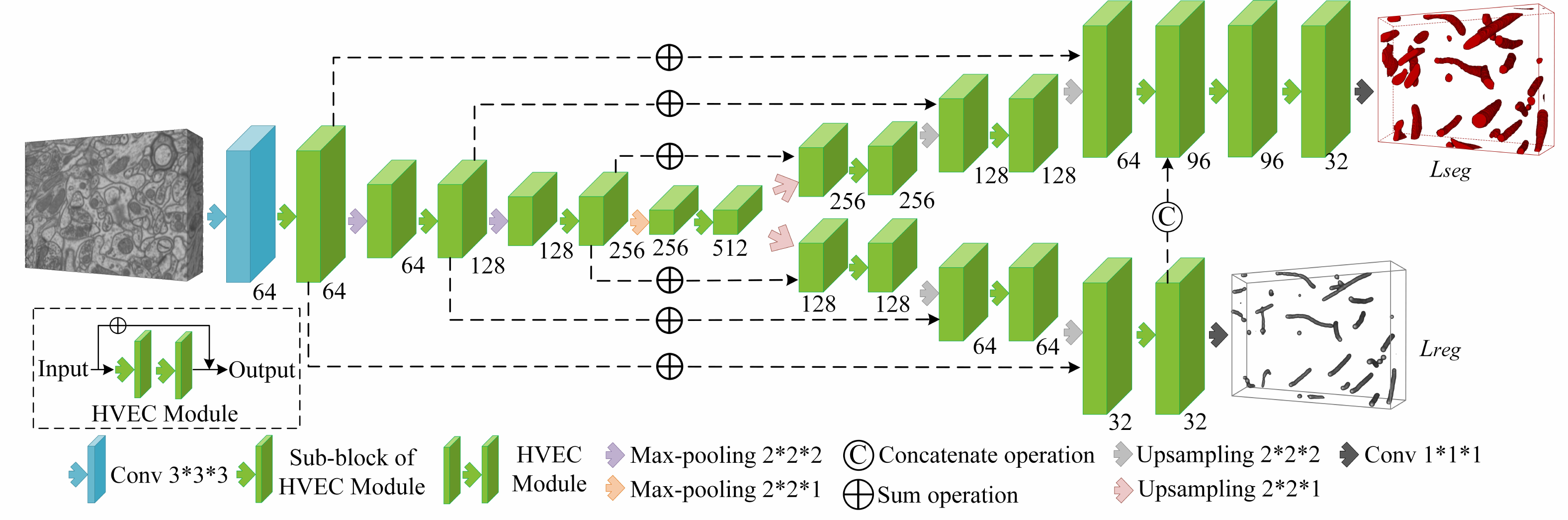}}
 \medskip
\end{minipage}
\caption{The architecture of our proposed EM-Net. The EM-Net is a multi-task learning based model which integrates two closely related tasks in a single network. The bottom branch performs mitochondria centerline detection to account for shape information for both enhancing the continuities and removing false  positives from the segmentation results.}
\label{fig:net_architecture}
\end{figure*}

\section{Introduction}
\label{sec:intro}

Mitochondria segmentation is a essential step for neuroscientists to investigate neuronal structures using electron microscopy (EM) images. Nowadays, with the advancement of EM imaging technology, massive unlabeled EM images can be easily acquired. Unfortunately, manually delineation such unprecedented scale of high-resolution data is time-consuming, tedious and also limited reproducibility. However, most existing automated segmentation methods rely on massive labeled data which are not available in most circumstances. Consequently, a fully automated mitochondria segmentation algorithm that requires limited annotated data is valuable  to help neurologist analyze EM images.

In virtue of the irregular shape variance, shift size of mitochondria and its complex background in EM images, fully automated mitochondrial segmentation has proven to be challenging. Previous studies on mitochondria segmentation mainly focus on designing hand-crafted features and classical machine learning classifiers. For instance, Lucchi \cite{lucchi2013learning}  employed ray descriptors (i.e. shape  cues) and histogram features to segment mitochondria. In a recent work \cite{peng2019mitochondria}, a class of local patch pattern (LPP) is introduced to encode contextual features to enhance mitochondria segmentation accuracy.
 Although these methods achieved promising performance, their performance usually soon reach saturation with the increasing size of data, due to the limited representability of hand-crafted features and small capacity of shallow models.

With the advent of deep fully convolutional networks (e.g, U-Net in 2D \cite{ronneberger2015u} or 3D \cite{cciccek20163d}), the performance of medical image segmentation has been pushed to a new level. Casser \textit{et al.} \cite{casser2018fast} proposed a modified 2D U-Net for mitochondria segmentation with on-the-fly data augmentation and reduced down-sampling stages. Their method, however, did not take full advantage of 3D spatial context. Xiao \textit{et al.} \cite{xiao2018automatic} proposed a fully residual U-Net using 3D convolution kernel and deep supervision,
which has achieved the state-of-the-art  performance. Despite of the superior performance of 3D convolutional networks, they  bring  a significant increase  of model parameters, thus need massive labeled data for good performance.
 Cheng \textit{et al.} \cite{cheng2017volume} introduced a light-weight 3D residual convolutional network by approximating a 3D convolution  with three rank-one (i.e.,1D) kernels. Modified residual connections with  the stochastic downsampling, are used to achieve feature-level data augmentation. Although using this factorized 3D kernel can greatly reduce model parameters, the 1D convolutions have  limited ability to capture crucial 3D spatial context.

To address the shortcomings mentioned above, we propose a centerline-aware multi-task network named EM-Net, which  not only can utilize intrinsic shape cues to regularize the segmentation, but also is light-weight using only 2D convolutions. Specifically, we integrate two closely related tasks, i.e. segmentation and centerline detection into a single network. The objective is to take account of the geometrical information of mitochondria represented by centerline to help improve the generalization performance and robustness of segmentation, especially when scarce annotated training samples are available. Moreover, a novel hierarchical view-ensemble convolution (HVEC) based module is introduced to reduce learning parameters and computation cost, and ensemble information on multiple 2D views of a 3D volume. The experiments showed that the proposed approach yielded superior results even with quite limited training data.

\section{METHOD}
\label{sec:method}

Our method comprises a main task for semantic 3D segmentation as well as an auxiliary centerline detection task to account for shape information of mitochondria. We formulate the segmentation task as a voxel-wise labeling problem, and the task of centerline detection as a regression problem. The ground truth for  regression is  the proximity score map generated using the centerline annotations for the mitochondria.

The architecture of our proposed EM-Net is depicted in Fig. \ref{fig:net_architecture}. More specifically, the EM-Net consists of one shared encoder  and two task-specific decoders, where each decoder path accounts for one task. The total loss of our model is as:
\begin{equation}
L_{total} = \lambda L_{seg} + (1-\lambda) L_{reg},
\end{equation}
where $\lambda$ is a trade-off parameter, and set to 0.7 in all experiments. Other than using 3D convolutions, we propose a novel \textit{hierarchical view-ensemble convolution} (HVEC) as building blocks for both encoder and decoder. Each decoder/encoder has three down-sampling/up-sampling stages.
  At decoding stage, low resolution feature maps generated by the encoder are progressively restored to input patch size. We replace deconvolutional operation with trilinear interpolation operation to restore the feature maps without additional parameters.

Following the idea of U-Net \cite{ronneberger2015u, cciccek20163d} we use skip-connections to integrate low level cues from layers of the encoder to the corresponding layers of decoder. Instead of using concatenation as U-Net, we use sum operation to achieve long-range residual learning. In fact, concatenation-based operation inevitably increases the number of feature channels.

\begin{figure}[htb]

\begin{minipage}[b]{1.0\linewidth}
  \centering
  \centerline{\includegraphics[width=0.9\textwidth]{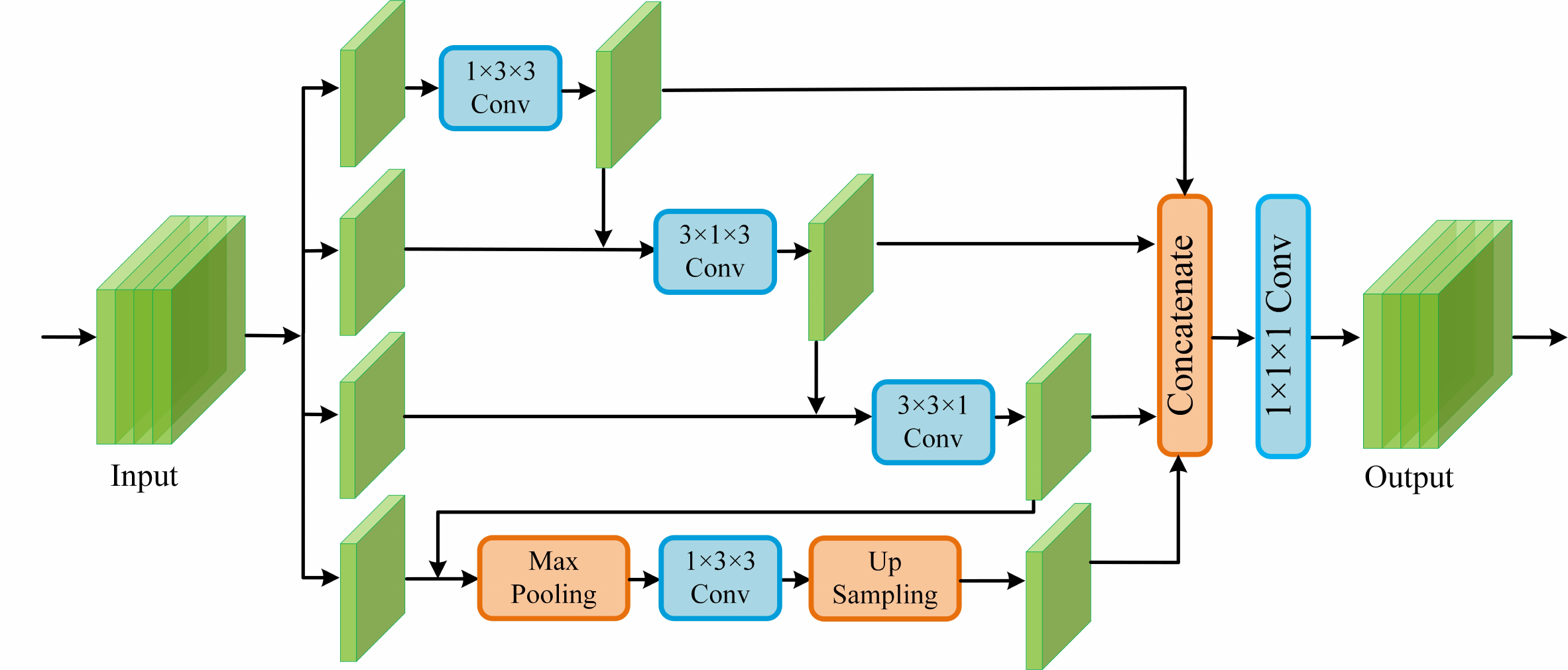}}
\end{minipage}
\caption{One sub-block of HVEC module. A HVEC module consists of two sub-blocks with short connection cross them for residual learning.}
\label{fig:HVEC}
\end{figure}

\textbf{HVEC module.}  As shown in Fig. 2, we firstly partition the input features into 4 groups and each group produces its own outputs, which are finally fused with concatenation.  Information cross different branches are then integrated with 1$\times$1$\times$1 convolution.  Compared with a standard 3D convolution layer attempting to simultaneously learn filters in all the three spatial dimensions and one channel dimension, the proposed factorized scheme is more parameter efficient. Note that factorized kernels have been explored in many works, such as depth-wise separable kernel in Inception \cite{szegedy2016rethinking}, group separable kernel in \cite{gao2019res2net}. However, our HVEC module is different from these methods in four aspects: 1) instead of conducting 3D convolutions on each feature group, we perform different 2D convolutions (i.e. 1$\times$3$\times$3, 3$\times$1$\times$3, 3$\times$3$\times$1) on the first three groups to encode information of three separable orthogonal views of a 3D volume; 2) on the fourth  group of features, 1$\times$3$\times$3 convolutions are performed on down-sampled features to capture context information at large scale on a focal view; 3) to capture multi-scale contexts and multiple fields-of-view, the four branches are convoluted in serial fashion, and the feature maps convoluted by previous branch are also added to the next branch as input, resulting hierarchical connections; 4) a whole HVEC module consists of two sub-blocks, and shortcut connections across two HVEC sub-blocks are to reformulate it as learning residual function in medium range. In this way, multi-scale and long-range multi-view context information, which are critical to the complicated EM image segmentation, can be encoded in a single module with significantly reduced parameters.

\textbf{Centerline detection task.} Instead of classifying each voxel as  centerline or  background, we formulate  it as a regression problem \cite{kainz2015you}. The ground truth for the centerline  regression is the proximity score map that is a distance transform function with peak at mitochondria centerlines and zeros on the background. Formally, it is defined as,
\begin{equation}
D(x) =
\begin{cases}
e^{\alpha(1-\frac {D_{C}(x) }{d_{M}})} - 1,  & \text{if $D_{C}(x) < d_{M}$} \\
0, & \text{otherwise}
\end{cases}
\end{equation}
where $\alpha$ and ${d_{M}}$ are  hyper parameters to govern the shape of the exponential function, and $D_{C}(x)$ represents the  minimum Euclidean distance between a voxel to the mitochondria centerline. The proximity score map $D$ is utilized as the supervision  to train our centerline detection sub-network. At the end of the detection path, the final feature maps is followed by a Sigmod layer to get the predicted proximity score map $D_{out}$. Minimizing mean squared error  $L_{reg}$ between $D_{out}$ and $D$ is utilized to accomplish the centerline detection task.

It is noticeable, however, that the number of layers and channels for centerline detection  is smaller than that for segmentation. Intuitively, the reasoning for this design is two fold: a) there are more smaller proportion of voxels taking  positive proximity scores than that of voxels taking positive labels. Thus, a deeper and wider network may result in over-fitting; b) increasing the number of feature channels will consume more GPU memory, and in turn decreases the size of network input, which is valuable for accurate segmentation.

\textbf{Segmentation task.} Compared with the detection path, the segmentation path has one more HVEC module, as shown in Fig. \ref{fig:net_architecture}. Moreover, the feature maps produced by the last HVEC module in the detection path will be concatenated to segmentation path. The goal is to take full advantage of the location and shape cues contained in the detection path.
We utilize Jaccard loss function for segmentation which is robust to the severe class imbalance in EM data. It is written as,
\begin{equation}
L_{seg} = 1.0 -  \frac { \sum_{i} P_i \cdot Y_i   }{\sum_{i} P_i + \sum_{i} Y_i - \sum_{i} P_i \cdot Y_i + \epsilon},
\end{equation}
where $P_i$ is the prediction for voxel $i$,  $Y_{i}$ is the ground truth, and   $\epsilon$ (e.g., $10^{-5}$) is to prevent dividing by zero.

\begin{figure}[t]
\begin{minipage}[b]{1.0\linewidth}
  \centerline{\includegraphics[width=1\textwidth]{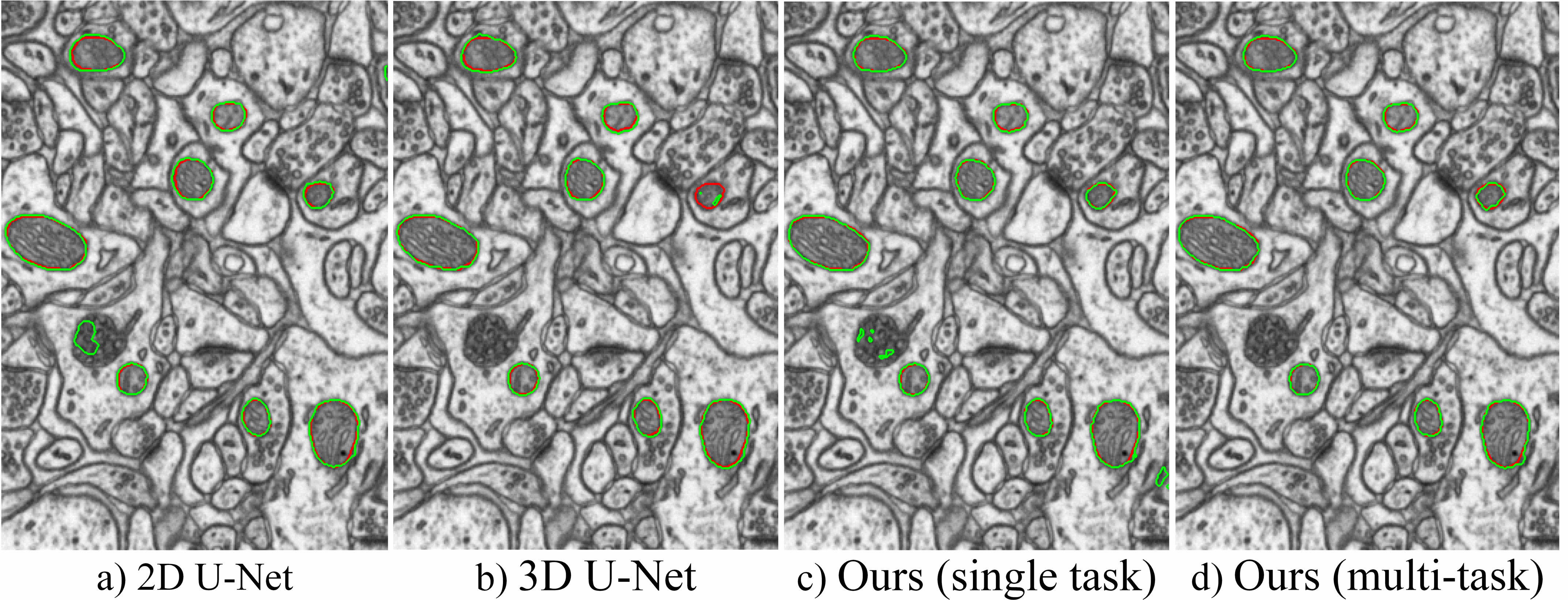}}
\end{minipage}
\caption{The visual comparison of different  methods.}
\label{fig:res}
\end{figure}


\begin{table}
\caption{Comparison of different methods for mitochondria segmentation on the FIB-SEM dataset. }
\centering
\label{tab:1}      
\begin{tabular}{lcc}
\hline\noalign{\smallskip}
Methods & Dice ($\%$) & JAC ($\%$)  \\
\noalign{\smallskip}\hline\noalign{\smallskip}
Lucchi \cite{lucchi2013learning} & 86.0 & 75.5\\
Peng \cite{peng2019mitochondria} & 90.9 & 83.3 \\
2D U-Net \cite{ronneberger2015u} & 91.5 & 84.4\\
Cheng (2D) \cite{cheng2017volume} & 92.8 & 86.5\\
3D U-Net \cite{cciccek20163d} &	93.5 &	87.8\\
Cheng(3D) \cite{cheng2017volume} &	94.1 &	88.9\\
Xiao \cite{xiao2018automatic} &	94.6 &	89.7\\
Ours (single task)	& 94.3 &	 89.2\\
Ours (multi-task)  & 	94.8	& 90.1\\
\noalign{\smallskip}\hline
\end{tabular}
\end{table}

\section{Results}
\label{sec:results}
For method evaluation, we use the EPFL dataset\footnote{https://cvlab.epfl.ch/data/em} with two annotated sub-stacks for training and testing, respectively. Each sub-stack  has a size of 165 $\times$768 $\times$ 1024,  with the voxel resolution of $5\times5\times5~nm$. These images are acquired by focused ion beam scanning EM (FIB-SEM), and taken from the CA1 hippocampus region of a mouse  brain. It is  the de facto standard benchmark for evaluation \cite{lucchi2013learning,peng2019mitochondria,cheng2017volume}.

We implement  models using PyTorch  with 1080Ti GPU. The model is optimized with Adam \cite{kingma2014adam}. The learning rate starts from 0.0001 and  a step-wise learning-rate decay scheme is used, where the step and decay rate are set to 15 and 0.9, respectively. We set $\alpha$=3, and $d_M$=15.  We use flipping, random transpose and  rotations for data augmentation  at training and  average  predictions of three rotated images
at testing.

We compare our method against state-of-the-art   methods  using hand-crafted features (\cite{lucchi2013learning}, \cite{peng2019mitochondria})  and deep learning(\cite{ronneberger2015u}, \cite{cciccek20163d}, \cite{cheng2017volume}, \cite{xiao2018automatic}).   For fair comparison, the same  data augmentation at training and testing stages as our method is used in the implementation of the compared deep learning methods.  Fig. \ref{fig:res} compares the results of our methods  with that of 2D U-Net and 3D U-Net. Visually, our methods have fewer false positives and discontinuities (missing regions).

  Quantitative comparisons illustrated in Table \ref{tab:1} are achieved with Dice and Jaccard-Index (JAC) measures. Overall, the algorithms based on deep learning significantly outperform traditional methods. Moreover, our method yields an accuracy of  90.1$\%$ in JAC, which is superior than most of other approaches.  The performance of both our method and 3D U-Net are superior than the 2D U-Net, which further confirms the importance of 3D spatial context. Meanwhile, it can be obviously seen that, even without centerline detection, our model still performs better than 3D U-Net, which validates the effectiveness of our HVEC module. Furthermore, thanks to the HVEC module, the parameters of our model are significantly less than 2D U-Net (3.2M vs. 31M) and 3D U-Net (3.2M vs. 19M). Note that we follow the setting in \cite{ronneberger2015u, cciccek20163d}  and use four down-sampling stages for 2D U-Net and three  down-sampling stages for 3D U-Net. By integration with the auxiliary centerline detection, our  method obtains a performance gain of 0.9\% in terms of JAC.  These validation results indicate that both our HVEC module and multi-task learning strategy are effective to improve performance.

\begin{figure}[t]
\begin{minipage}[b]{1.0\linewidth}
  \centerline{\includegraphics[width=0.5\textwidth]{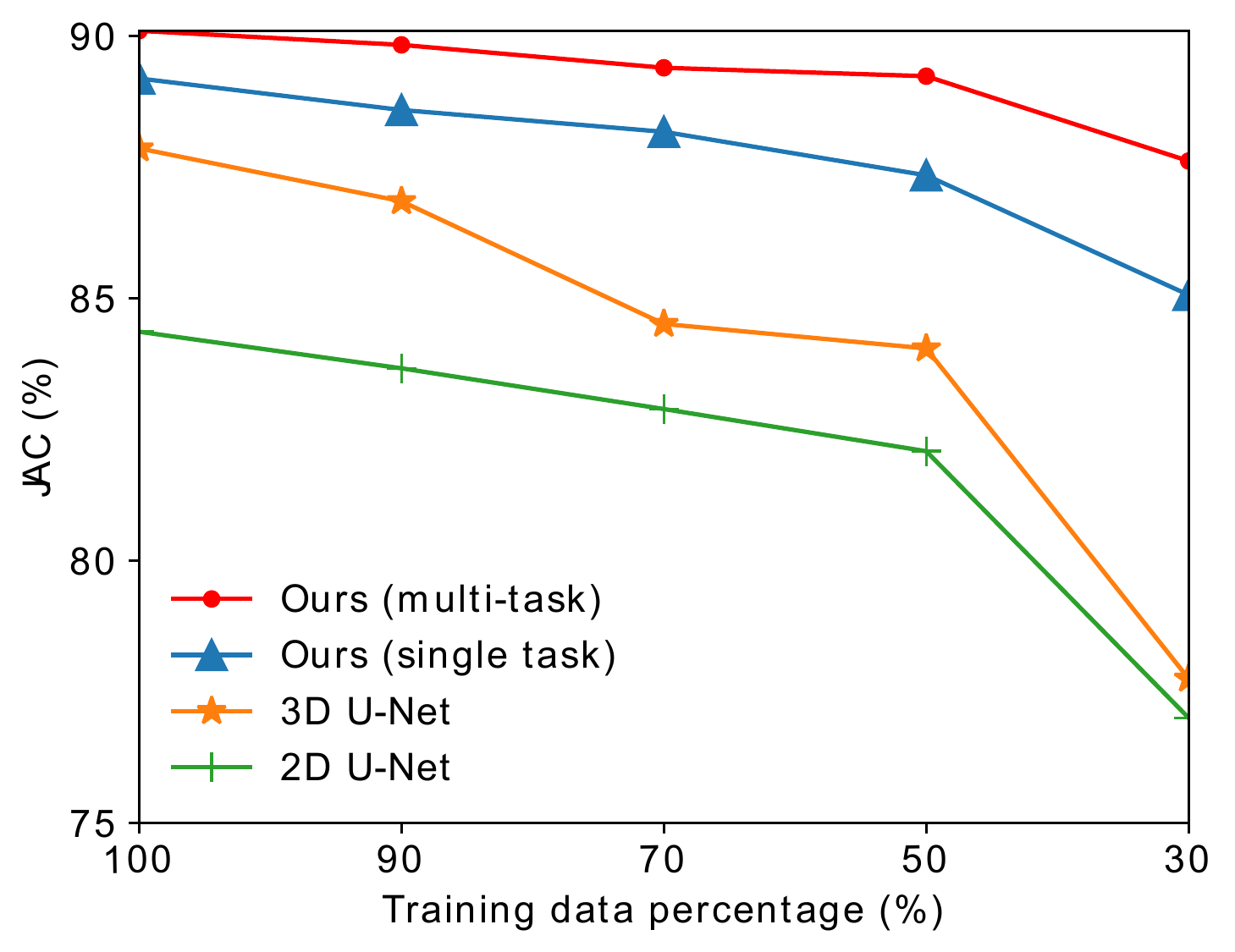}}
\end{minipage}
\caption{The comparative mitochondria segmentation performance on various fractions of training samples.}
\label{fig:data_decrease}
\end{figure}

To validate the robustness and  generalization ability of our model, we further investigate the segmentation performance of our  method under limited annotated data circumstances. Specifically, we gradually reduce the amount of training data, and test on the same testing dataset, the result of which is  depicted in Fig. \ref{fig:data_decrease}. Predictably, as the decrease of training data, all methods demonstrate  degenerated performance. Particularly, the 3D U-Net shows significant performance drop, and the gap between 3D U-Net and 2D U-Net narrows down quickly. However, our EM-Net and its single task variant are more robust to the reducing of training data, and is invariably higher than both 2D U-Net and 3D U-Net baseline methods. Note that, when the training data is decreased to only 30\%, all of the baseline methods have a sharp drop below 80$\%$ in performance. In contrast, our methods still can achieve sound performance (over 85\% in JAC), which confirms that the proposed EM-Net is an effective solution for mitochondria segmentation even in cases with scarce annotated training data.

\section{CONCLUSION}
\label{sec:majhead}
In this paper, we proposed a novel EM-Net for mitochondria segmentation. Specifically, we jointly perform  segmentation and centerline detection in a single network. To reduce the number of  parameters and substantial computational
cost, we introduce a novel hierarchical view-ensemble convolution (HVEC), a simple
alternative of 3D convolution to learn 3D spatial contexts using more efficient 2D convolutions. Validation and comparison  results on the EPFL public benchmark showed that, our method can not only  significantly reduce learnable parameters, achieving  an efficient light-weight model, but also can obtain state-of-the-art performance.


\end{document}